\documentclass{article}
\PassOptionsToPackage{numbers, compress}{natbib}

\usepackage[final]{new_in_ML}

\usepackage[utf8]{inputenc} 
\usepackage[T1]{fontenc}    
\usepackage{hyperref}       
\usepackage{url}            
\usepackage{booktabs}       
\usepackage{amsfonts}       
\usepackage{nicefrac}       
\usepackage{microtype}      
\usepackage{xcolor}         
\usepackage{graphicx}
\usepackage{subfig}
\usepackage{multirow}
\usepackage{balance}

\title{NutritionVerse-Real: An Open Access Manually Collected 2D Food Scene Dataset for Dietary Intake Estimation}

\author{
Chi-en Amy Tai$^{1}$ \quad Saeejith Nair$^{1}$ \quad Olivia Markham$^{1}$ \quad Matthew Keller$^{1}$ \\ \quad \textbf{Yifan Wu}$^{1}$ \quad \textbf{Yuhao Chen}$^{1}$  \quad \textbf{Alexander Wong}$^{1,2}$\\
$^1$University of Waterloo, Waterloo, Ontario, Canada\\
$^2$Waterloo Artificial Intelligence Institute, Waterloo, Ontario, Canada\\
{\tt\small {\{amy.tai, smnair, olivia.markham, m6keller, yifan.wu1,}} \\ {{\tt\small yuhao.chen1, a28wong\}}@uwaterloo.ca}
}

\begin{document}

\maketitle

\begin{abstract}
Dietary intake estimation plays a crucial role in understanding the nutritional habits of individuals and populations, aiding in the prevention and management of diet-related health issues. Accurate estimation requires comprehensive datasets of food scenes, including images, segmentation masks, and accompanying dietary intake metadata. In this paper, we introduce NutritionVerse-Real, an open access manually collected 2D food scene dataset for dietary intake estimation with 889 images of 251 distinct dishes and 45 unique food types. The NutritionVerse-Real dataset was created by manually collecting images of food scenes in real life, measuring the weight of every ingredient and computing the associated dietary content of each dish using the ingredient weights and nutritional information from the food packaging or the Canada Nutrient File. Segmentation masks were then generated through human labelling of the images. We provide further analysis on the data diversity to highlight potential biases when using this data to develop models for dietary intake estimation. NutritionVerse-Real is publicly available at \url{https://www.kaggle.com/datasets/nutritionverse/nutritionverse-real} as part of an open initiative to accelerate machine learning for dietary sensing. 
\end{abstract}

\section{Introduction}
Dietary intake estimation plays a crucial role in understanding the nutritional habits of individuals and populations, aiding in the prevention and management of diet-related health issues~\cite{malnutrition-qol}. Unfortunately, there is substantial bias with conventional dietary intake methods like food frequency questionnaires, food diaries, and 24-hour recall~\cite{automated-dietary-recall, freedman2014pooled, freedman2015pooled}. Subsequently, there has been immense focus on alternative automated approaches for dietary intake estimation that leverage emerging work in computer vision and machine learning~\cite{food-recog-promise, 10.1145/3347448.3357172}. However, accurate estimation requires comprehensive datasets of food scenes, including images, segmentation masks, and accompanying dietary intake metadata. 

\begin{figure}[!ht]
    \centering
    \subfloat[Example Dish 1]{\includegraphics[width=.22\linewidth]{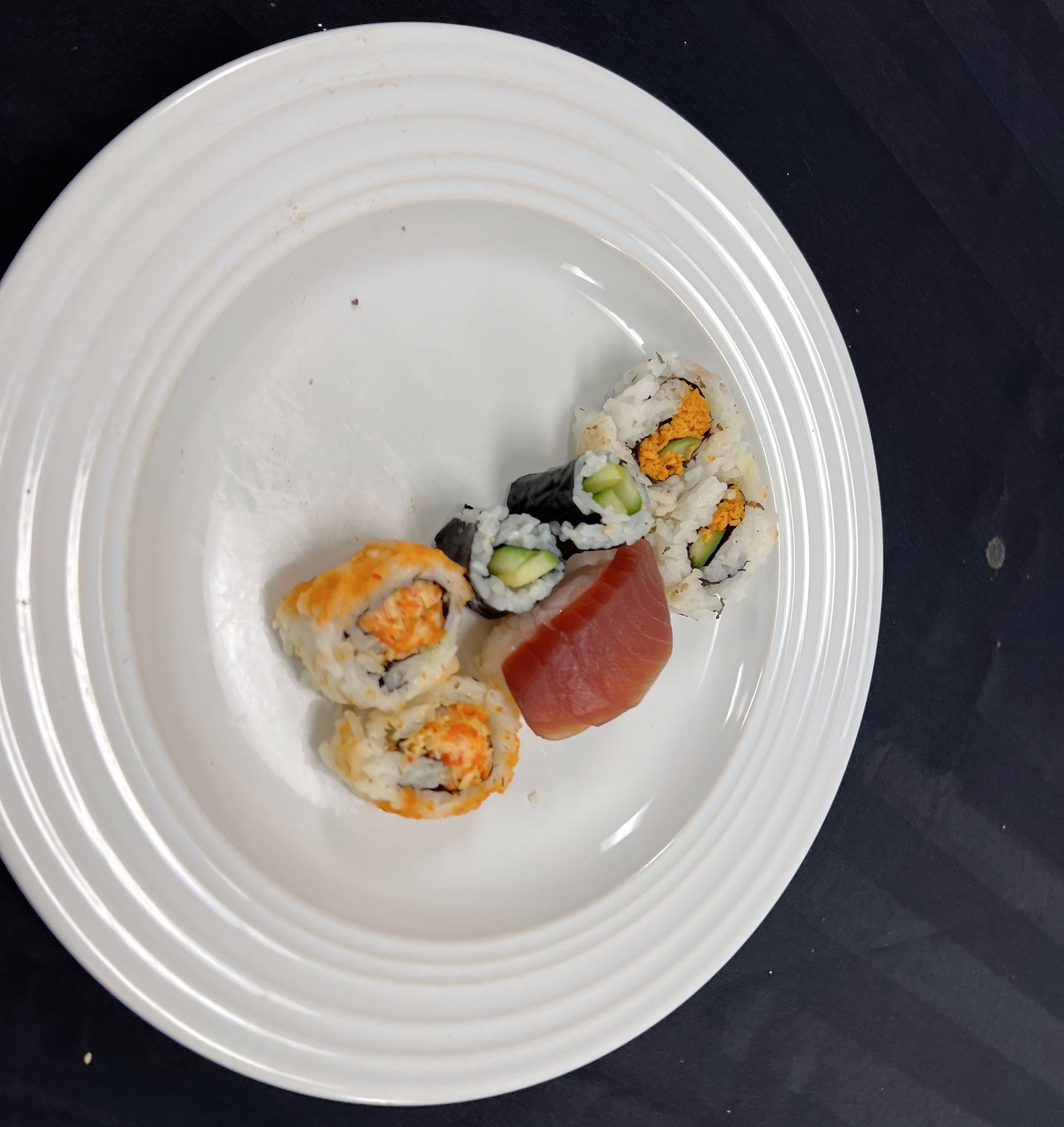}}
    \hfil
    \subfloat[Example Dish 2]{\includegraphics[width=.22\linewidth]{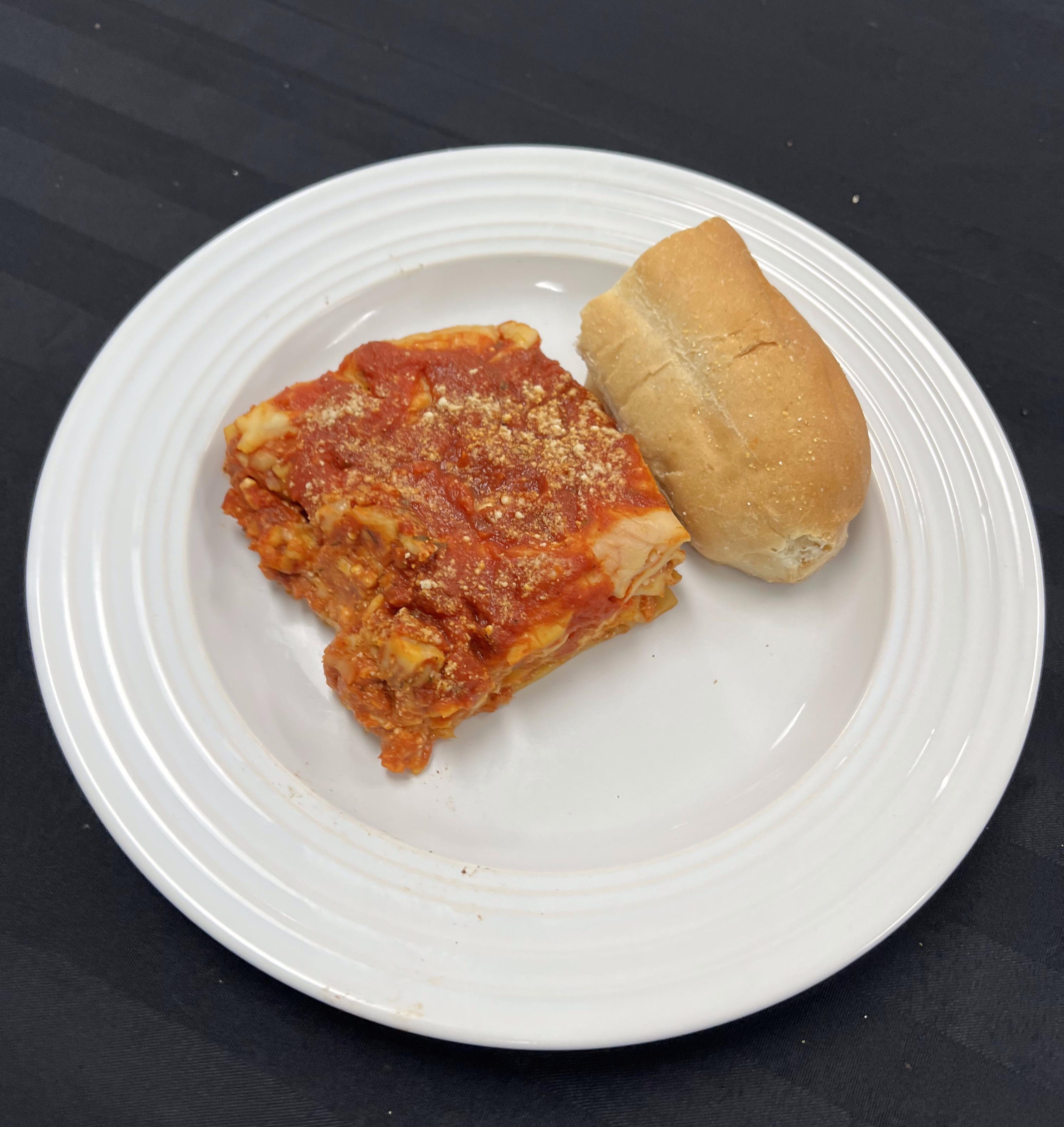}}
    \hfil
    \subfloat[Example Dish 3]{\includegraphics[width=.22\linewidth]{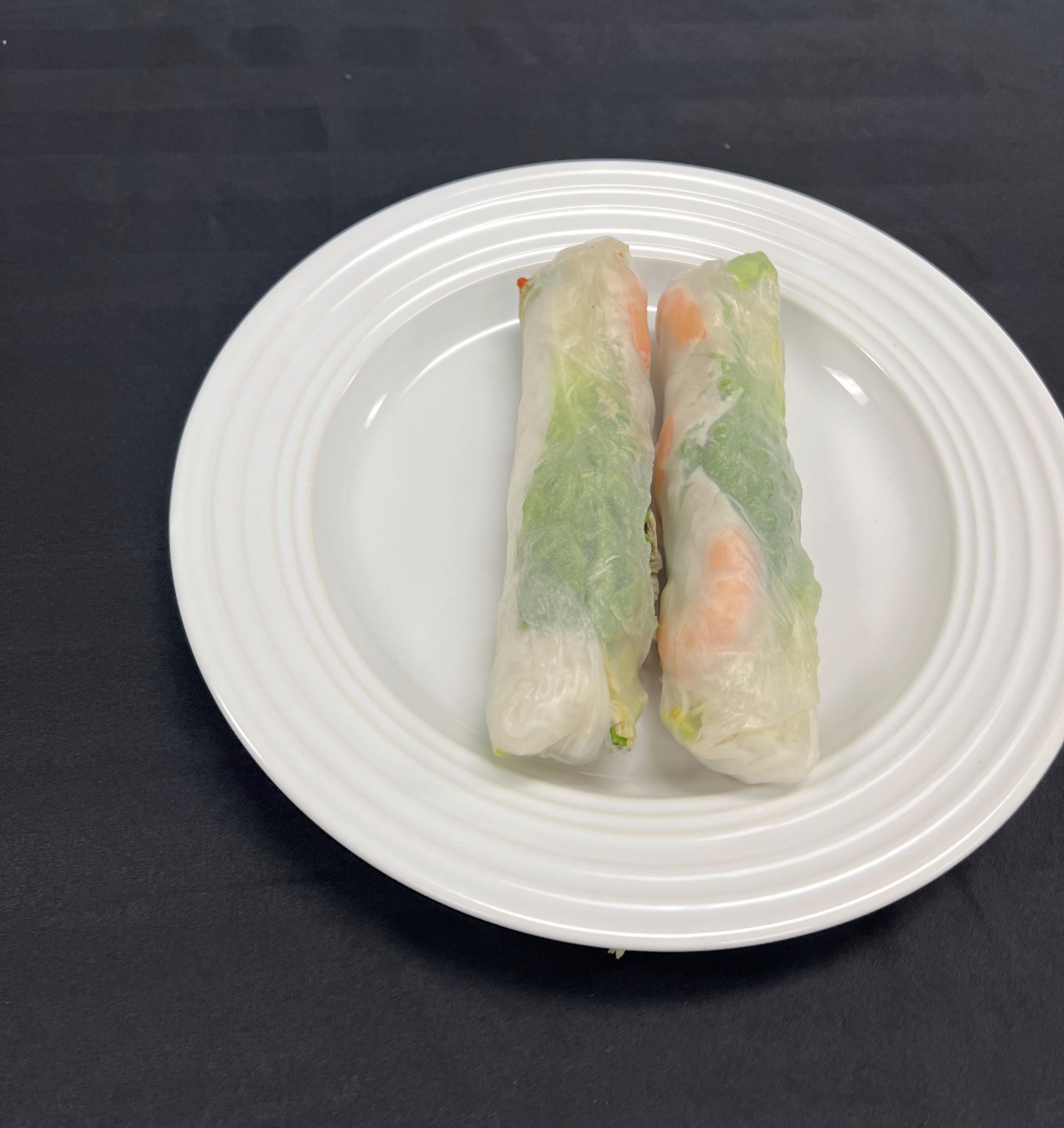}}
    \hfil
    \subfloat[Example Dish 4]{\includegraphics[width=.22\linewidth]{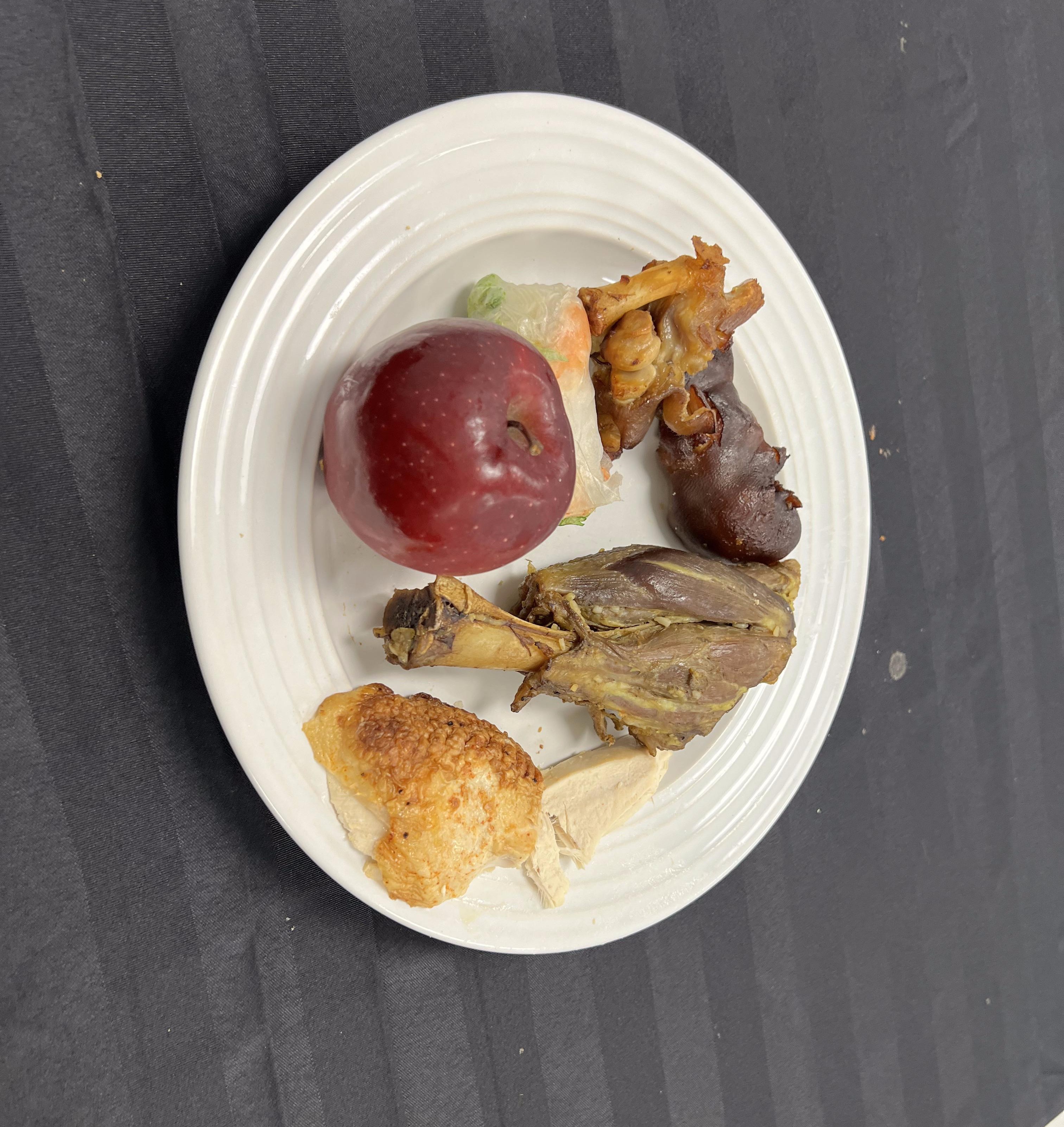}}
    \caption{Example dishes from NutritionVerse-Real dataset.}
    \label{fig:example-dishes}
\end{figure}

In this paper, we introduce NutritionVerse-Real, an open access manually collected 2D food scene dataset for dietary intake estimation. Sample images can be seen in Figure~\ref{fig:example-dishes}. NutritionVerse-Real is publicly available at \url{https://www.kaggle.com/datasets/nutritionverse/nutritionverse-real} as part of an open initiative to accelerate machine learning for dietary sensing. 

\section{Methodology} 
\begin{figure}[!ht]
    \centering
    \subfloat[Example Dish 1]{\includegraphics[width=.47\linewidth]{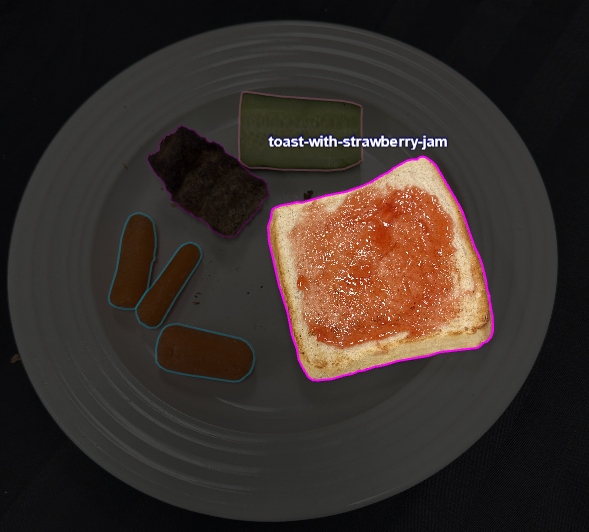}}
    \hfil
    \subfloat[Example Dish 2]{\includegraphics[width=.51\linewidth]{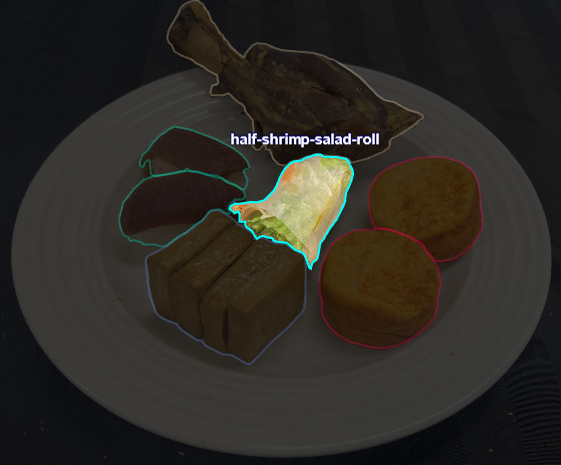}}
    \caption{Examples of the segmentation mask for scenes labelled using Roboflow in the NutritionVerse-Real dataset.}
    \label{fig:example-label-dishes}
\end{figure}

The NutritionVerse-Real dataset was created by manually collecting images of food scenes in real life. For each dish, ten images were collected from random camera angles using an iPhone 13 Pro Max. The weight of every ingredient in the dish was measured using a food scale and the nutritional values for each ingredient was obtained from the packaging or the Government of Canada's Canada Nutrient File~\cite{canada-nutrient-file}. The weight and average nutritional value was then used to compute the nutritional values (e.g., protein, calories, fat, carbohydrate) of the ingredients, which were summed to get the total nutritional value of the dish. Segmentation masks were then generated through human labelling of the images using Roboflow~\cite{dwyer2022roboflow}. Four of the ten images for each scene were selected at random for manual mask labelling. Examples of the segmentation mask for scenes labelled using Roboflow in the NutritionVerse-Real dataset is shown in Figure~\ref{fig:example-label-dishes}.

\section{Results and Discussion}
\begin{figure}[!ht]
    \centering
    \includegraphics[width=0.5\linewidth]{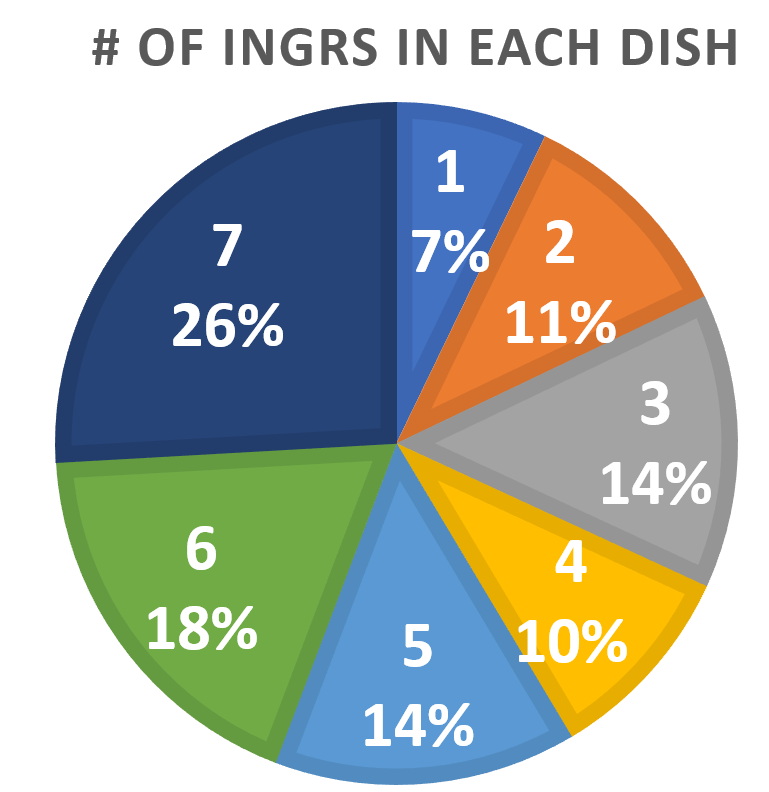}
    \caption{Distribution of number of ingredients in a dish.}
    \label{fig:distribution-num-ingrs}
\end{figure}

NutritionVerse-Real contains 889 2D images of 251 distinct dishes and 45 unique food types. In terms of data diversity, each food item appears at least once in 18 dishes on average, highlighting relatively fair distribution of food types in the dataset. As seen in Figure~\ref{fig:distribution-num-ingrs}, each dish has between 1 to 7 ingredients with 26\% of dishes having 7 ingredients in the dish. 

\begin{figure}[!t]
    \centering
    \subfloat[Calories (\% of DV for Female)]{\includegraphics[width=.5\linewidth]{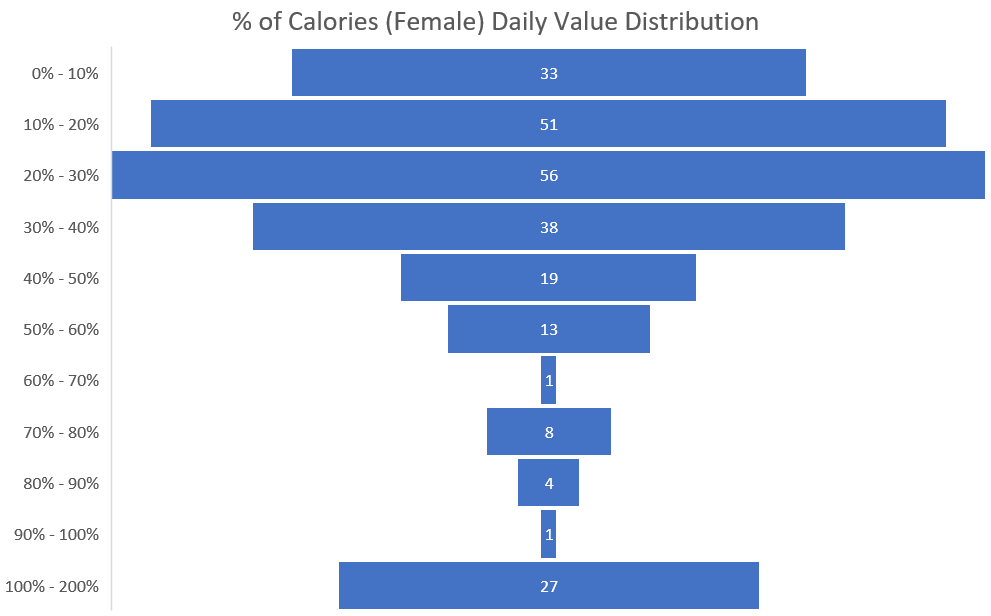}}
    \hfil
    \subfloat[Calories (\% of DV for Male)]{\includegraphics[width=.5\linewidth]{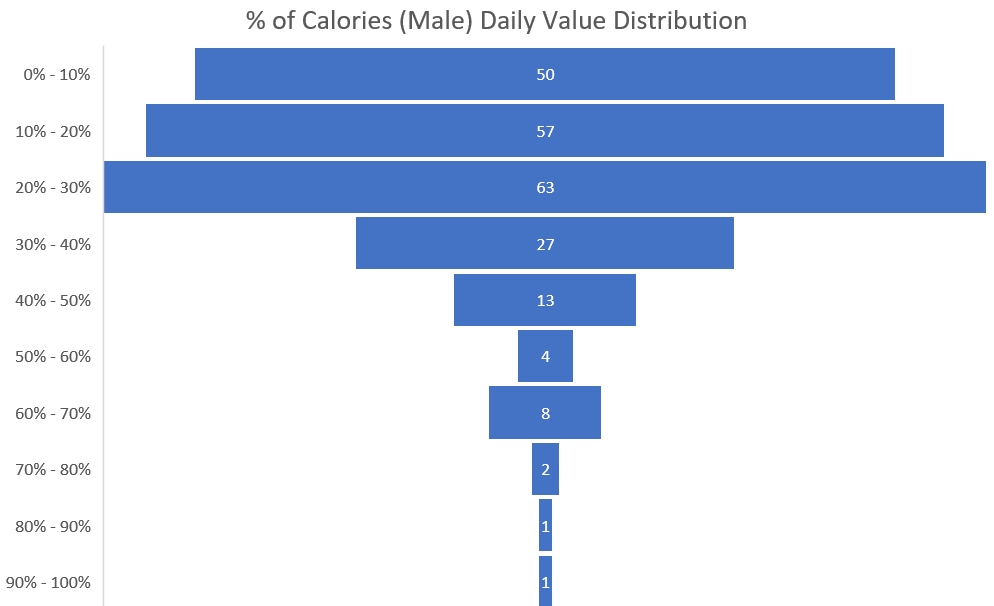}}
    \hfil
    \subfloat[Fat (\% of DV)]{\includegraphics[width=.5\linewidth]{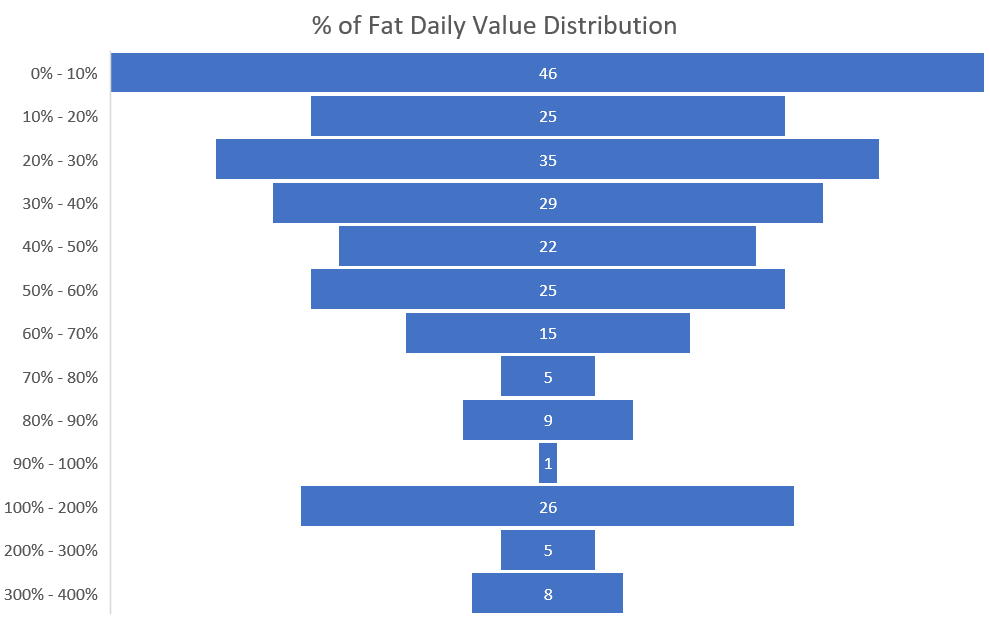}}
    \hfil
    \subfloat[Carbohydrates (\% of DV)]{\includegraphics[width=.5\linewidth]{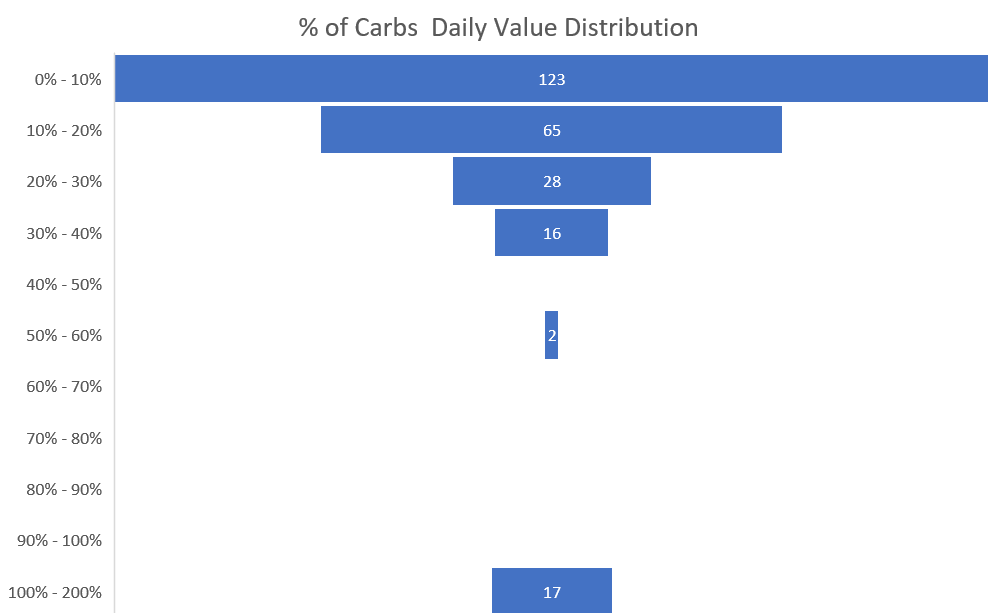}}
    \caption{Distribution of the dataset across various macronutrients as a percent of the daily value (DV) obtained from~\cite{percent-daily-value,osilla2018calories}.}
    \label{fig:macro-distribution}
\end{figure}

On average, each dish has a total food weight of 406 g, 830 calories, 38.2 g of fat, 59.9 g of carbohydrates, 64 g of protein, 0.17 mg of calcium, 0.01 mg of iron, 0.11 mg of magnesium, and 0.91 mg of potassium. As depicted in Figure~\ref{fig:macro-distribution}, the distribution of the total dish nutritional values within the dataset as a percentage of the daily value is fairly well distributed for calories (female) and fat, but skewed towards the lower range of the nutrient category for calories (male) and carbohydrates. This skewed distribution is not merely a limitation but rather reflects the realistic challenges often encountered with dataset distributions in the real world. This situation highlights the critical need for models that can robustly handle such data imbalances. While there is a potential bias towards predicting lower nutrient values, acknowledging and addressing this issue presents a vital opportunity for further research.

\begin{ack}
This work was supported by the National Research Council Canada (NRC) through the Aging in Place (AiP) Challenge Program, project number AiP-006. The authors thank undergraduate research assistants Anthony Susevski and Samridhi Gupta. 
\end{ack}

{
\small

\bibliography{new_in_ML}

\begin{thebibliography}{10}

\bibitem{malnutrition-qol}
Heather~H. Keller, Truls Østbye, and Goy Richard.
\newblock Nutritional risk predicts quality of life in elderly community-living canadians.
\newblock {\em The Journals of Gerontology: Series A}, 59(1):M68–M74, 2004.

\bibitem{automated-dietary-recall}
Amy~F. Subar, Sharon~I. Kirkpatrick, Beth Mittl, Thea~Palmer Zimmerman, Frances~E. Thompson, Christopher Bingley, Gordon Willis, Noemi~G. Islam, Tom Baranowski, Suzanne McNutt, and Nancy Potischman.
\newblock The automated self-administered 24-hour dietary recall (asa24): A resource for researchers, clinicians, and educators from the national cancer institute.
\newblock {\em Journal of the Academy of Nutrition and Dietetics}, 112(8):1134--1137, 2012.

\bibitem{freedman2014pooled}
Laurence~S Freedman, John~M Commins, James~E Moler, Lenore Arab, David~J Baer, Victor Kipnis, Douglas Midthune, Alanna~J Moshfegh, Marian~L Neuhouser, Ross~L Prentice, et~al.
\newblock Pooled results from 5 validation studies of dietary self-report instruments using recovery biomarkers for energy and protein intake.
\newblock {\em American journal of epidemiology}, 180(2):172--188, 2014.

\bibitem{freedman2015pooled}
Laurence~S Freedman, John~M Commins, James~E Moler, Walter Willett, Lesley~F Tinker, Amy~F Subar, Donna Spiegelman, Donna Rhodes, Nancy Potischman, Marian~L Neuhouser, et~al.
\newblock Pooled results from 5 validation studies of dietary self-report instruments using recovery biomarkers for potassium and sodium intake.
\newblock {\em American journal of epidemiology}, 181(7):473--487, 2015.

\bibitem{food-recog-promise}
Gianluigi Ciocca, Paolo Napoletano, and Raimondo Schettini.
\newblock Food recognition: A new dataset, experiments, and results.
\newblock {\em IEEE Journal of Biomedical and Health Informatics}, 21(3):588--598, 2017.

\bibitem{10.1145/3347448.3357172}
Yoshikazu Ando, Takumi Ege, Jaehyeong Cho, and Keiji Yanai.
\newblock Depthcaloriecam: A mobile application for volume-based foodcalorie estimation using depth cameras.
\newblock In {\em Proceedings of the 5th International Workshop on Multimedia Assisted Dietary Management}, MADiMa '19, page 76–81, New York, NY, USA, 2019. Association for Computing Machinery.

\bibitem{canada-nutrient-file}
Government of~Canada.
\newblock Canadian nutrient file (cnf) - search by food, 2022.

\bibitem{dwyer2022roboflow}
B.~Dwyer, J.~Nelson, J.~Solawetz, et~al.
\newblock Roboflow (version 1.0) [software], 2022.
\newblock computer vision.

\bibitem{percent-daily-value}
Government of~Canada.
\newblock Percent daily value, 2019.

\bibitem{osilla2018calories}
Eva~V Osilla, Anthony~O Safadi, and Sandeep Sharma.
\newblock {\em Calories}.
\newblock StatPearls Publishing, Treasure Island (FL), 2018.

\end{thebibliography}
}

\end{document}